\documentclass[10pt,twocolumn,letterpaper]{article}

\usepackage[pagenumbers]{cvpr}      

\usepackage{bm}

\definecolor{cvprblue}{rgb}{0.21,0.49,0.74}
\usepackage[pagebackref,breaklinks,colorlinks,allcolors=cvprblue]{hyperref}
\usepackage{url}
\usepackage{graphicx}
\usepackage[table]{xcolor}
\usepackage{pifont}
\usepackage{makecell} 
\usepackage{multirow}

\def\paperID{*****} 
\def\confName{CVPR}
\def\confYear{2026}

\title{GRPO-Guard: Mitigating Implicit Over-Optimization in Flow Matching via Regulated Clipping}

\author{
Jing Wang$^{1,2\S}$, Jiajun Liang$^{2\ast}$, Jie Liu$^{3}$, Henglin Liu$^{2,4}$, Gongye Liu$^{2,5}$, Jun Zheng$^{1}$, Wanyuan Pang$^{6}$\\
Ao Ma$^{7}$, Zhenyu Xie$^{1}$, Xintao Wang$^{2}$, Meng Wang$^{2}$, Pengfei Wan$^{2}$, Xiaodan Liang$^{1\dag}$\\
$^1$Shenzhen Campus of Sun Yat-Sen University \quad 
$^2$Kling Team, Kuaishou Technology \\ \quad 
$^3$CUHK MMLab
$^4$Tsinghua University \quad 
$^5$HKUST \quad 
$^6$USTB \quad 
$^7$UCAS\\[3pt]
{\small \textit{\href{https://jingw193.github.io/GRPO-Guard/}{Project Page} $^\ast$Project Leader. $^\dag$Corresponding Authors. $^\S$Work Conducted During Internship.}}\\[-3pt]
{\fontsize{8pt}{9pt}\selectfont  \texttt{wangj977@mail2.sysu.edu.cn, liangjiajun@kuaishou.com}}
}

\begin{document}

\twocolumn[{
\renewcommand\twocolumn[1][]{#1}%
\maketitle
\vspace{-9mm}
\begin{center}
    \centering
    \includegraphics[width=1.0\linewidth]{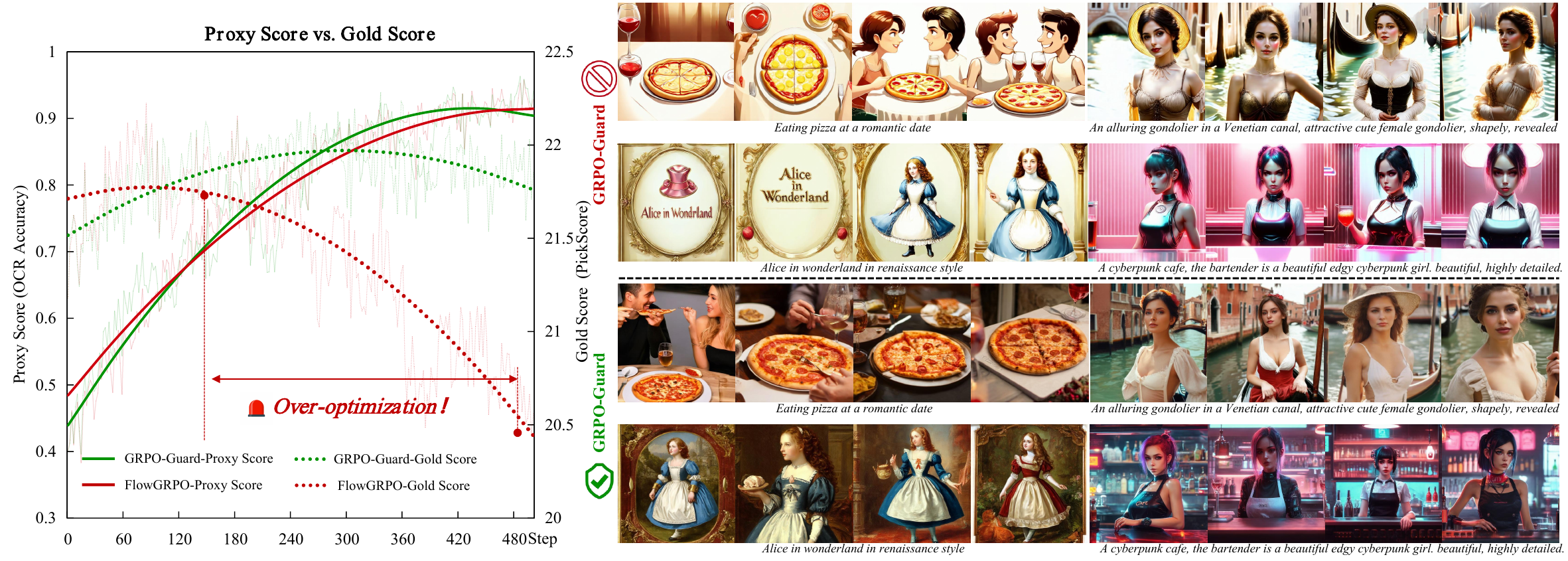}
    \captionof{figure}{\textbf{Comparison between FlowGRPO and GRPO-Guard under over-optimization}.
\textbf{Left}: The proxy score and gold score trends during training. As the proxy score increases, FlowGRPO rapidly enters an over-optimization phase, where the gold score continuously declines.
\textbf{Right}: A visual comparison between FlowGRPO and GRPO-Guard. Due to severe reward hacking, FlowGRPO suffers from a drastic degradation in diversity, detail richness, visual quality, and text-image consistency (bottom part). In contrast, GRPO-Guard maintains a stable gold score and high visual quality under a comparable proxy score, as shown in the upper part of the figure.}
    \label{figure1}
\end{center}%
}]

\begin{abstract}
Recently, GRPO-based reinforcement learning has shown remarkable progress in optimizing flow-matching models, effectively improving their alignment with task-specific rewards.
Within these frameworks, the policy update relies on importance-ratio clipping to constrain overconfident positive and negative gradients.
However, in practice, we observe a systematic shift in the importance-ratio distribution—its mean falls below 1 and its variance differs substantially across timesteps. This left-shifted and inconsistent distribution prevents positive-advantage samples from entering the clipped region, causing the mechanism to fail in constraining overconfident positive updates.
As a result, the policy model inevitably enters an \textbf{implicit over-optimization} stage—while the proxy reward continues to increase, essential metrics such as image quality and text–prompt alignment deteriorate sharply, ultimately making the learned policy impractical for real-world use.
To address this issue, we introduce \textbf{GRPO-Guard}, a simple yet effective enhancement to existing GRPO frameworks. Our method incorporates ratio normalization, which restores a balanced and step-consistent importance ratio, ensuring that PPO clipping properly constrains harmful updates across denoising timesteps. In addition, a gradient reweighting strategy equalizes policy gradients over noise conditions, preventing excessive updates from particular timestep regions. 
Together, these designs act as a regulated clipping mechanism, stabilizing optimization and substantially mitigating implicit over-optimization without relying on heavy KL regularization.
Extensive experiments on multiple diffusion backbones (e.g., SD3.5M, Flux.1-dev) and diverse proxy tasks demonstrate that GRPO-Guard significantly reduces over-optimization while maintaining or even improving generation quality. These results highlight GRPO-Guard as a robust and general solution for stable policy optimization in flow-matching models.
\end{abstract}    
\section{Introduction}
\label{sec:intro}

Recent advances in flow-based diffusion models~\cite{lipman2022flow, liu2022flow, peebles2023scalable} have led to remarkable progress in visual generation. State-of-the-art models such as Stable Diffusion 3.5 (SD3.5)~\cite{esser2024scaling}, Flux~\cite{flux2024}, and Qwen-Image~\cite{wu2025qwenimagetechnicalreport} achieve outstanding image synthesis quality, while Wan2.1~\cite{wan2025} and Kling~\cite{kling2024} extend this success to video generation. Building on the success of Group Relative Policy Optimization (GRPO~\cite{shao2024deepseekmath}) in large language models~\cite{guo2025deepseek, jaech2024openai}, recent works such as Flow-GRPO~\cite{liu2025flow} and DanceGRPO~\cite{xue2025dancegrpo} apply GRPO-style reinforcement learning to diffusion models, yielding notable improvements in aesthetic quality~\cite{kirstain2023pick, wu2023human}, instruction following~\cite{ghosh2023geneval}, and text rendering~\cite{chen2023textdiffuser}.

Within GRPO frameworks, the importance-ratio clipping mechanism serves primarily to stabilize training. By bounding policy updates when the new policy diverges excessively from the reference model, clipping suppresses gradient explosions and maintains controlled optimization across denoising timesteps. Ideally, the importance ratio should remain centered around 1, ensuring that positive and negative updates are symmetrically constrained—effectively truncating gradients from overconfident samples and preserving balance in policy learning.

However, our empirical analysis reveals that this stabilization mechanism fails to behave as intended in diffusion models. As shown in Figure~\ref{ratio_distribu}, the importance-ratio distribution exhibits a systematic bias: its mean consistently falls below 1, and its variance varies significantly across timesteps. This left-shifted and uneven distribution prevents positive-advantage samples from entering the clipped region, leaving overconfident positive updates largely unconstrained. As training progresses, the policy model gradually enters an over-optimization~\cite{miao2024inform} regime—the proxy reward continues to rise, while essential metrics such as image fidelity and text–prompt alignment degrade sharply, rendering the learned policy impractical for real-world use.
Furthermore, the variance inconsistency across timesteps exacerbates this imbalance: at high-noise steps, clipping is rarely activated, whereas at low-noise steps, it occurs excessively—leading to persistent over-optimization at high-noise steps. Taken together, the mean shift and variance disparity in the importance-ratio distribution amplify gradient imbalance across noise conditions, which we identify as the root cause of implicit reward hacking observed in Flow-GRPO (Figure~\ref{figure1}).

We trace this anomalous behavior to a fundamental design mismatch: diffusion models compute Gaussian probabilities, whereas LLMs rely on discrete token probabilities, yet FlowGRPO or DanceGRPO directly inherits the GRPO formulation without proper adaptation. To address this issue, we propose GRPO-Guard, a simple yet effective enhancement to existing GRPO frameworks. It introduces a ratio normalization (RatioNorm) procedure that standardizes the importance-ratio distribution at each denoising step, ensuring its mean remains close to one and its variance consistent across timesteps. This adjustment restores the clipping mechanism’s ability to truncate gradients from overconfident positive samples, mitigating imbalance-induced over-optimization and stabilizing policy learning.
Despite this correction, policy gradients still vary significantly across timesteps. Low-noise steps yield disproportionately large gradients, causing the policy model to overfit specific noise conditions while neglecting early-step exploration and diversity. This imbalance ultimately drives the model toward over-optimization concentrated at a single step. To alleviate this issue, we propose a gradient balancing strategy that treats gradients from all steps more uniformly, effectively mitigating over-optimization while providing modest performance gains.

As illustrated in Figure~\ref{ratio_distribu}, GRPO-Guard restores healthy ratio distributions and consistent clipping across timesteps, achieving fast convergence comparable to KL-free baselines while substantially reducing over-optimization. It consistently alleviates reward hacking across multiple GRPO variants (e.g., Flow-GRPO, DanceGRPO), diverse diffusion backbones (e.g., SD3.5-M, FLUX1.dev), and various proxy tasks (e.g., text rendering, GenEval, PickScore). This demonstrates the robustness, scalability, and general applicability of our approach to safe policy optimization in diffusion-based generation models.

\section{Related Works}

\subsection{Alignment for Large Language Models}
Recent years witness a shift from supervised fine-tuning to interactive, reinforcement-style alignment when adapting Large Language Models(LLMs)~\cite{achiam2023gpt} to human intent~\cite{shani2024multi, sun2024supervised}. Reinforcement Learning from Human Feedback (RLHF)~\cite{gao2023scaling} — which typically trains a reward model from pairwise human comparisons and then optimizes a policy using RL algorithms such as PPO~\cite{schulman2017proximal} — becomes a standard pipeline for this purpose~\cite{ouyang2022training, christiano2017deep, bai2025qwen2}. However, PPO-based RLHF pipelines are often computationally intensive and sensitive to reward-model inaccuracies, which has motivated the development of more stable and efficient alternatives. One such direction is Direct Preference Optimization (DPO)~\cite{rafailov2023direct}, which bypasses explicit reinforcement learning by directly optimizing model parameters on human preference pairs, achieving similar alignment effects with reduced complexity. More recently, Group Relative Policy  Optimization(GRPO) methods have already been adopted in production-scale LLM alignment flows~\cite{jaech2024openai, guo2025deepseek}, demonstrating that group-relative updates can yield stable improvements in instruction following and preference alignment.

\subsection{RL for Diffusion and Flow Models.}
Diffusion and flow-matching models~\cite{ho2020denoising, song2020denoising, rombach2022high, lipman2022flow} decompose the process of visual generation into iterative denoising steps, revolutionizing the field of visual synthesis and achieving remarkable results in both image and video generation. Building on the success of reinforcement learning (RL) algorithms in Large Language Models (LLMs), similar optimization paradigms—such as PPO~\cite{schulman2017proximal, black2023training} and DPO~\cite{wallace2024diffusion}—have been effectively transferred to diffusion models, enabling preference alignment and improved task-specific performance. Following this trend, Flow-GRPO~\cite{liu2025flow} and DanceGRPO~\cite{xue2025dancegrpo} integrate GRPO-style policy updates into flow-matching models, transforming deterministic ODE sampling into stochastic SDE formulations to introduce exploration noise for group-based optimization. More recently, MixGRPO~\cite{li2025mixgrpo} proposes a hybrid ODE–SDE sampling strategy that significantly improves training efficiency while maintaining comparable generation quality. Meanwhile, Flow-CPS~\cite{wang2025coefficients} identifies a critical issue in the SDE sampling process used by Flow-GRPO and DanceGRPO—namely, the inconsistency of noise coefficients across timesteps—which leads to excessive residual noise and inaccurate reward estimation. To address this, Flow-CPS introduces a noise-consistent SDE sampling scheme that accelerates GRPO optimization by improving reward accuracy.
In parallel, TempFlowGRPO~\cite{he2025tempflow} and G$^2$RPO~\cite{zhou2025text} address the reward sparsity and inaccuracy caused by assigning a single global reward to multi-step SDE trajectories. 
Most existing methods focus on improving policy optimization efficiency but overlook a critical issue—over-optimization, which severely degrades visual quality. In this work, we conduct an in-depth analysis of this problem and propose an effective solution.

\subsection{Reward Over-optimization.}
Reward over-optimization~\cite{gao2023scaling, moskovitz2023confronting}, also referred to as reward hacking~\cite{skalse2022defining, miao2024inform}, poses a significant challenge in reinforcement learning for diffusion and flow models, arising from the limitations of imperfect proxy reward models~\cite{liu2025improving, xu2023imagereward, wang2025unified} (RMs) for human or task-specific preferences. In practice, optimizing a learned proxy RM often improves its corresponding proxy metric, but alignment with the true objective—such as perceptual quality or human-evaluated preference—typically holds only for a short period, after which further optimization can degrade generation quality, as illustrated in Figure ~\ref{figure1}.

To mitigate this issue, common strategies include regularizing policy updates with a heavy KL-divergence penalty~\cite{fan2023dpok, liu2025flow} toward a supervised fine-tuned policy. KL regularization helps mitigate over-optimization by reducing drift from the reference policy, but it can also slow the improvement of both proxy scores and true-performance metrics, potentially leading to degraded overall performance. Clipping importance ratios~\cite{schulman2017proximal} further constrains updates from overly confident positive and negative samples, preventing harmful updates and stabilizing policy optimization, thereby reducing the risk of entering an over-optimization phase. Additionally, scaling up reward models~\cite{gao2023scaling, wu2025rewarddance}, using ensembles~\cite{coste2023reward, eisenstein2023helping}, or composing RMs from multiple perspectives can further reduce overfitting to a single proxy, although at significant computational cost. Early stopping~\cite{black2023training} and monitoring generation quality provide additional safeguards against excessive reward exploitation, but they may also halt training prematurely, potentially leaving the policy under-optimized.

However, in flow-matching models, the inherent bias in the importance ratio causes the clipping mechanism to fail to function as intended, allowing overly confident positive updates to pass unchecked and driving the policy into an over-optimization regime. In this work, we analyze this phenomenon in depth and propose methods to mitigate implicit over-optimization, thereby restoring stable and reliable policy updates.
\section{Method}

\subsection{Preliminary}

\paragraph{Flow Matching:}assumes that $x_1 \sim X_1$ is a Gaussian noise sample and $x_0 \sim X_0$ is a sample drawn from the real data distribution. The Rectified Flow formulation defines the noisy sample $x_t$ as
\begin{equation}
    x_t = (1 - t)x_0 + t x_1,
\end{equation}

A Transformer-based model $v_\theta$ is trained to predict the velocity field $v = x_1 - x_0$. The training objective of Flow Matching is to minimize the expected squared error between the predicted and true velocities:

\begin{equation}
L(\theta) = \mathbb{E}_{t, x_0 \sim X_0, x_1 \sim X_1 }[\|v - v_\theta (x_t, t) \|^2].
\end{equation}

\paragraph{Flow-GRPO and DanceGRPO:} During the reinforcement learning (RL) stage, Flow-GRPO and DanceGRPO introduce stochasticity into the sampling process by converting the ODE-based deterministic flow used in Flow Matching into a stochastic differential equation (SDE) formulation. The SDE sampling process in Flow-GRPO and DanceGRPO can be expressed as:
\begin{align}
\label{x_tdt}
    &x_{t+ dt} = \\ &\underbrace{x_t + \left[ v_\theta(x_t, t) + \frac{\sigma_t^2}{2t}(x_t + (1 - t)v_\theta(x_t, t))\right]dt}_{\mu_\theta(x_t, t)} + \sigma_t \sqrt{dt} \epsilon \notag
\end{align}
where $\bm{\epsilon} \sim \mathcal{N}(0,\bm{I})$. In Flow-GRPO, the noise level $\sigma_t$ is defined as $\sigma_t = \eta \sqrt{\frac{t}{1-t}}$. In contrast, DanceGRPO adopts a constant noise level $\sigma = \eta$.

Subsequently, given the same conditioning input $c$, a group of diverse samples ${x_0^i}_{i = 1}^G$ is generated through the SDE sampling process. Each sample is evaluated by the reward model, which assigns a scalar score $R(x_0^i)$. The group-relative advantage is then computed as:
\begin{align}
\hat{A_t^i} = \frac{R(x_0^i) - \text{mean}({R(x_0^i)}_{i=1}^G)}{\text{std}({R(x_0^i)}_{i=1}^G))}
\end{align}
The GRPO algorithm then optimizes the policy model by minimizing the following objective:
\begin{align}
    \mathcal{J}_{\text{policy}}(\theta)
= \frac{1}{G} \sum_{i=1}^{G} 
\frac{1}{T} \sum_{t=0}^{T-1}
\Big(
\min\!\big(r_t^{i}(\theta)\, \hat{A}_t^{i},\; \notag\\
\text{clip}(r_t^{i}(\theta),\, 1 - \epsilon,\, 1 + \epsilon)\, \hat{A}_t^{i}\big)
\Big),
\end{align}

where $r_t^{i}(\theta) = \frac{ p_{\theta}(x_{t-1}^i|x_t^i, \bm{c})}{p_{\theta_{old}}(x_{t-1}^i|x_t^i, \bm{c})}$. For Flow-GRPO, an additional KL penalty $D_{KL}(\pi_\theta || \pi_{ref})$ is introduced to mitigate reward hacking, constraining the policy model to stay close to the reference flow model. DanceGRPO enforces that the initial random noise for samples within the same group $\{x_1^i\}_{i=1}^G$, remains identical, ensuring that all generated variations originate from the same starting point.

\subsection{Analysis and Solution}
In this section, we first analyze the importance ratio in Flow-GRPO and DanceGRPO, highlighting the underlying causes of its inherent distributional anomalies and their role in inducing implicit reward hacking. We then introduce the RatioNorm method, which corrects the importance-ratio distribution, regulates the clipping mechanism, and mitigates reward hacking. Finally, we propose a gradient reweighting strategy to prevent single-step gradients from dominating optimization, thereby alleviating over-optimization under specific noise conditions.

\begin{figure*}[h]
    \includegraphics[width=1.0\linewidth]{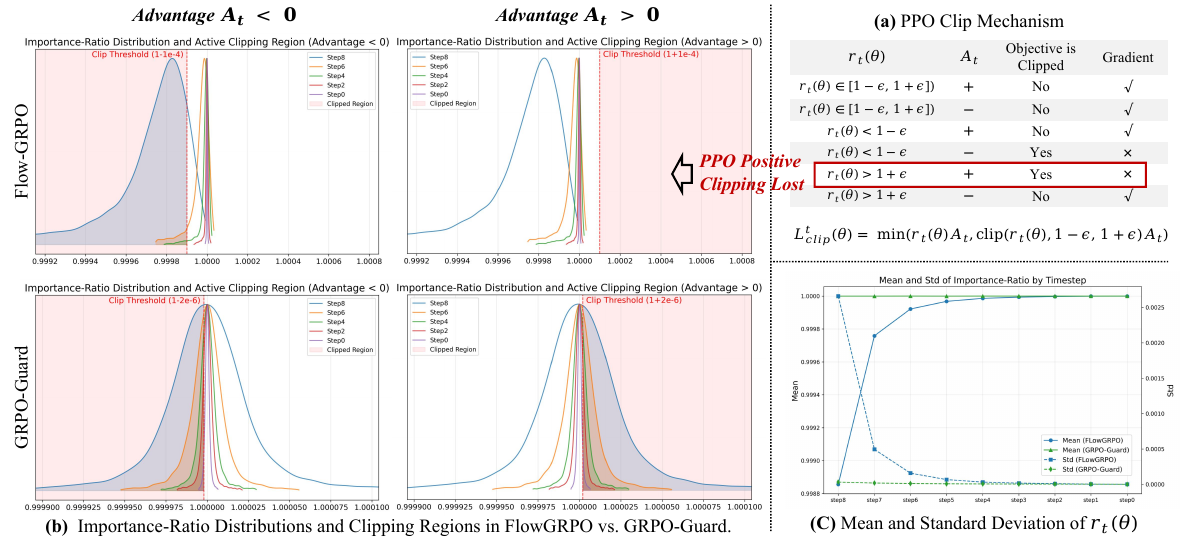}
    \caption{\textbf{Comparison of $r(\theta)$ distributions between FlowGRPO and GRPO-Guard across timesteps.} (a) Ideally, the ratio distribution should have a mean near 1 and stable variance across timesteps to ensure effective clipping. (b) Under FlowGRPO, the distribution exhibits a leftward mean shift and increasing variance at low-noise timesteps, causing the clipping mechanism to fail—particularly for trajectories with positive advantages. In contrast, GRPO-Guard with RatioNorm preserves a balanced mean and consistent variance (c), enabling proper clipping and stable policy updates across all timesteps.}
    \label{ratio_distribu}
\end{figure*}

\subsubsection{Inherent Distributional Anomalies}
Ideally, the importance ratio should stay centered around 1, so that positive and negative updates are symmetrically constrained—truncating gradients from overconfident samples and maintaining balanced policy learning, as illustrated in Figure~\ref{ratio_distribu}(a). However, in diffusion models, the importance-ratio distribution often exhibits abnormal shifts, causing the clipping mechanism for positive samples to fail. We analyze this issue in detail in the following section.

In flow matching, the log-probability $\log p_\theta(x_{t-1}|x_t, \bm{c})$ under the policy model $\theta$, is computed using the Gaussian probability formula:
\begin{equation}
\log p_{\theta}(x_{t-1}|x_t, \bm{c}) = -\frac{\|x_{t-1} - \mu_\theta(x_t,t)\|^2}{2\sigma_t^2  dt} - C_t, 
\end{equation}
where $x_{t-1} = \mu_{\theta_{old}}(x_t,t) + \sigma_t \sqrt{dt} \cdot \bm{\epsilon}$, $\bm{\epsilon} \sim \mathcal{N}(0,\bm{I})$ and $C_t$ is a constant. Consequently, we can derive the expression for the log-importance ratio $\log r_t(\theta)$ as follows:
\begin{align} 
\log r_t(\theta) &= \log p_{\theta}(x_{t-1}|x_t, \bm{c}) - \log p_{\theta_{old}}(x_{t-1}|x_t, \bm{c}) \notag\\
&= -\frac{\|\mu_{\theta_{old}}(x_t,t) - \mu_\theta(x_t,t) + \sigma_t \sqrt{dt} \cdot \bm{\epsilon} \|^2}{2\sigma_t^2dt} \notag\\ &+ \frac{\|\mu_{\theta_{old}}(x_t,t) - \mu_{\theta_{old}}(x_t,t) + \sigma_t \sqrt{dt} \cdot \bm{\epsilon} \|^2}{2\sigma_t^2dt} \notag\\
&= -\frac{\|\Delta \mu_\theta + \sigma_t \sqrt{dt} \cdot \bm{\epsilon} \|^2}{2\sigma_t^2dt} + \frac{\| \bm{\epsilon} \|^2}{2} \notag\\
&= - \frac{\|\Delta \mu_\theta\|^2}{2\sigma_t^2dt} - \frac{\Delta \mu_\theta \cdot \bm{\epsilon}}{\sigma_t \sqrt{dt}}
\end{align}

For simplicity, we denote $\Delta \mu_\theta =\mu_{\theta_{old}}(x_t,t) - \mu_{\theta}(x_t,t)$.
Since $\bm{\epsilon} \sim \mathcal{N}(0,\bm{I})$, we illustrate the derivation with a one-dimensional Gaussian (without loss of generality). Then we have $\mathbb{E}_{\bm{\epsilon} \sim \mathcal{N}(0,\bm{I})}\left[\log r_t(\theta) \right] = - \frac{\|\Delta \mu_\theta\|^2}{2\sigma_t^2\Delta t}$.

This analysis reveals a key distinction from LLMs: unlike discrete token probabilities in language models, diffusion models compute Gaussian state transition probabilities. The resulting quadratic term introduces a timestep-dependent negative bias in the log-importance ratio, as illustrated in Figure~\ref{ratio_distribu}(b). Because the expected ratios are generally below 1, samples with positive advantage rarely exceed the upper clipping bound. Consequently, gradients from overconfident positive predictions are largely retained, while those from negative samples are more heavily constrained, making the policy susceptible to over-optimization.

Additionally, the variance of the importance ratio depends on denoising scheduler parameters such as $\sigma_t$ and $dt$, causing it to differ substantially across timesteps. This variance inconsistency further amplifies clipping imbalance: at low-noise steps, the clipping threshold is frequently exceeded, while at high-noise steps, it is rarely triggered, ultimately driving the policy toward step-specific over-optimization.

\subsubsection{Regulated Clipping}

A straightforward approach would be to design a dedicated clipping range for each timestep. However, this requires tuning a large number of hyperparameters and performing extensive experiments to identify near-optimal ranges for a specific model and task. To simplify this process and quickly reduce timestep-dependent differences in the mean and variance of the importance-ratio distribution, we instead standardize $\log r_t(\theta)$. This normalization shifts the mean toward zero and removes the influence of denoising scheduler parameters, while preserving the sign and relative magnitude of $\Delta \mu_\theta$, thereby maintaining the semantic content of the ratios.
Specifically, the operation is defined mathematically as:
\begin{align}
\log \hat{r_t}(\theta) &= \sigma_t \sqrt{dt} (\log r_t(\theta) + \frac{\|\Delta \mu_\theta\|^2}{2\sigma_t^2 dt}) \notag \\
&= - \Delta \mu_\theta \cdot \epsilon
\end{align}
After normalization using the above formula, the distribution of the ratios is illustrated in the Figure \ref{ratio_distribu}(b): the mean approaches zero. As a result, the upper and lower clipping bounds can now function effectively, improving the training stability of Flow-GRPO. 

However, we observe that the log ratios still exhibit substantial variance differences across timesteps, which leads to uneven clipping when a single clip range is applied. Specifically, in high-noise steps, gradients are rarely clipped, and samples with positive advantage and large ratios retain their full gradient contribution, so over-optimization still occurs. We analyze that this variance discrepancy is mainly caused by coefficients related to the noise term, which are inherently correlated with timestep characteristics in diffusion models, as shown in the Figure \ref{ratio_distribu}(c). To address this, we remove the influence of the noise coefficients, thereby reducing the variance across timesteps and mitigating the reward hacking phenomenon.

Since the log ratios undergoes multiplication and addition operations related to the denoising step, it introduces additional effects on the policy gradient. We will further analyze this phenomenon in detail below.

\vspace{-5pt}
\subsubsection{Gradient Analysis}
\begin{figure*}[h]
    \includegraphics[width=1.0\linewidth]{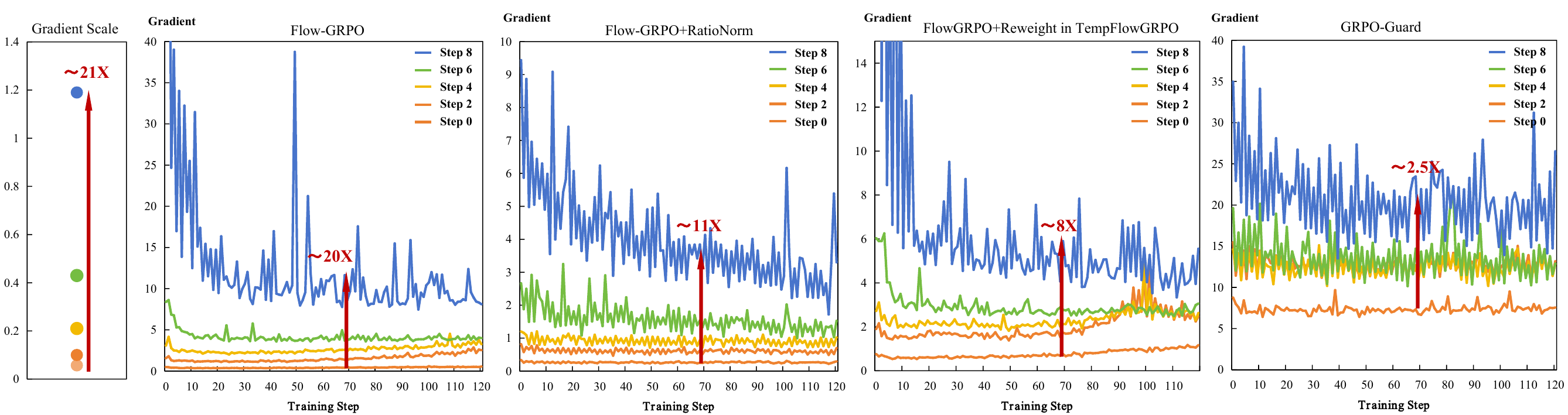}
    \caption{\textbf{Gradient magnitude differences across timesteps}. In FlowGRPO, gradient magnitudes vary by roughly 20× across timesteps, reflecting the large differences in gradient scale. GRPO-Guard substantially reduces this imbalance, limiting the variation to about 2.5× and preventing over-optimization under any single noise condition.}
    \label{grad_fig}
\end{figure*}

First, we revisit the formulation of the policy gradient in FlowGRPO. Consider the sampling and training process of a policy model, whose policy gradient in Flow-GRPO or Dance-GRPO can be formulated as follows. For simplicity, we omit the clipping operation, the minimization term, and the KL-penalty component:
\begin{align}
& \nabla_{\theta} \mathcal{J}(\theta)
= \sum_{t=0}^{T-1}\hat{A_t}\, \nabla_{\theta}r_t(\theta)   = \sum_{t=0}^{T-1}\hat{A_t}\,r_t(\theta)\nabla_{\theta} \log r_t(\theta) \notag\\
&=  \sum_{t=0}^{T-1} 
\mathbb{E}_{\epsilon \sim \mathcal{N}(0, \bm{I})}
\left[
\hat{A_t} \,r_t(\theta) \nabla_{\theta} \log p_{\theta}(x_{t-1}| x_t) 
\right] \\
&= \sum_{t=0}^{T-1} 
\mathbb{E}_{\epsilon \sim \mathcal{N}(0, \bm{I})}
\left[
 \frac{\Delta \mu_\theta + \sigma_t \sqrt{dt} \cdot \epsilon}{\sigma_t^2dt}\hat{A_t}\,r_t(\theta)\nabla_{\theta}\mu_\theta(x_t | t)
\right] \notag
\end{align}
According to Eq. \ref{x_tdt}, we have $\nabla_{\theta}\mu_\theta(x_t | t) = (1 + \frac{\sigma_t^2(1 - t)}{2t})dt \nabla_{\theta}v_\theta(x_t | t)$. 
In FlowGRPO, since $\sigma_t = \eta \sqrt{\frac{t}{1 - t}}$, the coefficient $(1 + \frac{\sigma_t^2 (1 - t)}{2t}) = 1 + \frac{\eta^2}{2}$ remains approximately constant across timesteps. We thus simplify it as a constant term $\beta$. Therefore,
\begin{align}
&\nabla_{\theta} \mathcal{J}(\theta)
= \\ &\sum_{t=0}^{T-1} 
\mathbb{E}_{\epsilon \sim \mathcal{N}(0, \bm{I})}
\left[
\beta \frac{\Delta \mu_\theta + \sigma_t \sqrt{dt} \, \epsilon}{\sigma_t^2}  \,\hat{A_t}\,r_t(\theta)\nabla_{\theta}v_\theta(x_t | t)
\right]. \notag
\end{align}

where $\beta \frac{\Delta \mu_\theta + \sigma_t \sqrt{dt} \, \epsilon}{\sigma_t^2}$ are defined as the gradient scale that is independent of the advantage term. 
Then we empirically analyze and visualize the policy gradients and corresponding gradient scales across different denoising steps in Flow-GRPO.
As shown in Figure~\ref{grad_fig}, both exhibit a strong correlation and demonstrate a consistent trend of increasing gradient magnitude as the noise level decreases.

This result aligns with the observations in TempFlowGRPO~\cite{he2025tempflow}, which addresses this issue through a noise-aware reweighting strategy. Specifically, it introduces a reweighting coefficient of ($\sigma_t \sqrt{dt}$), adjusting the gradient scaling to ($\beta\,dt \,\epsilon$) for the on-policy case and ($\beta \,\frac{\sqrt{dt}\Delta \mu_\theta + \sigma_t dt,\epsilon}{\sigma_t}$) for the off-policy case, thereby improving the optimization efficiency of the policy model.

Subsequently, we observe that after applying RatioNorm, the policy gradient becomes less sensitive to these gradient scaling factors. The detailed formulation is as follows:

\begin{align}
& \nabla_{\theta} \mathcal{J}(\theta)
= \sum_{t=0}^{T-1}\hat{A_t}\, \nabla_{\theta} \hat{r_t}(\theta)   = \sum_{t=0}^{T-1}\hat{A_t}\,\hat{r_t}(\theta)\nabla_{\theta} \log \hat{r_t}(\theta) \notag\\
&=  \sum_{t=0}^{T-1} 
\mathbb{E}_{\epsilon \sim \mathcal{N}(0, \bm{I})}
\left[
\epsilon \, \hat{A_t} \,\hat{r_t} (\theta) \,  \nabla_{\theta}\mu_\theta(x_t | t)
\right]  \notag\\
&=  \sum_{t=0}^{T-1} 
\mathbb{E}_{\epsilon \sim \mathcal{N}(0, \bm{I})}
\left[
\beta \, dt\, \epsilon \, \hat{A_t}\, \hat{r_t}(\theta)  \nabla_{\theta}v_\theta(x_t | t)
\right]
\end{align}
Here, both $r_t(\theta)$ and $\hat{r_t}(\theta)$ typically lie within the range of $[1 - 1e^{-3}, 1 + 1e^{-3}]$, making their direct influence on the gradient negligible. After applying RatioNorm, the policy gradient becomes more accurate by removing the interference from factors such as $\Delta \mu_\theta$ and $\sigma_t$. As shown in Figure~\ref{grad_fig}, the gradient imbalance is alleviated, and the gradient scale approaches $\beta \,dt \,\epsilon$, resembling the on-policy gradient reweighting in TempFlowGRPO.

However, the gradient scale is still influenced by the timestep-dependent coefficient $dt$. We argue that this leads certain steps to dominate the optimization process, as the policy update tends to focus on a single noise condition within the entire sampling trajectory, thereby increasing the risk of over-optimization. As illustrated in Figure~\ref{fig:ablation}, although the reweighting strategy of TempFlowGRPO accelerates optimization, it also makes the policy more susceptible to over-optimization.

To further alleviate the over-optimization issue, we incorporate a reweighting factor $\delta = 1 / dt$ into the policy loss. As illustrated in Figure~\ref{grad_fig}, this adjustment effectively normalizes the gradient magnitudes across different timesteps, leading to a more stable optimization process.
It is worth noting that for DanceGRPO, $\sigma_t = \eta$, so the coefficient becomes $\beta = 1 + \frac{\eta^2(1 - t)}{2t}$. Consequently, the reweighting factor in DanceGRPO is defined as $\delta = \beta / dt$.

The final form of our policy loss is expressed as follows:

\begin{align}
    \mathcal{J}_{\text{policy}}(\theta)
= \frac{1}{G} \sum_{i=1}^{G} 
\frac{1}{T} \sum_{t=0}^{T-1}
\Big( \textcolor{red}{\text{$\delta$}}
\min\!\big(\textcolor{red}{\text{$\hat{r_t^{i}}(\theta)$}}\, \hat{A}_t^{i},\; \notag\\
\text{clip}(\textcolor{red}{\text{$\hat{r_t^{i}}(\theta)$}},\, 1 - \epsilon,\, 1 + \epsilon)\, \hat{A}_t^{i}\big)
\Big)
\end{align}

\begin{figure*}
  \centering
  \begin{subfigure}{0.33\linewidth}
    \includegraphics[width=1.0\linewidth]{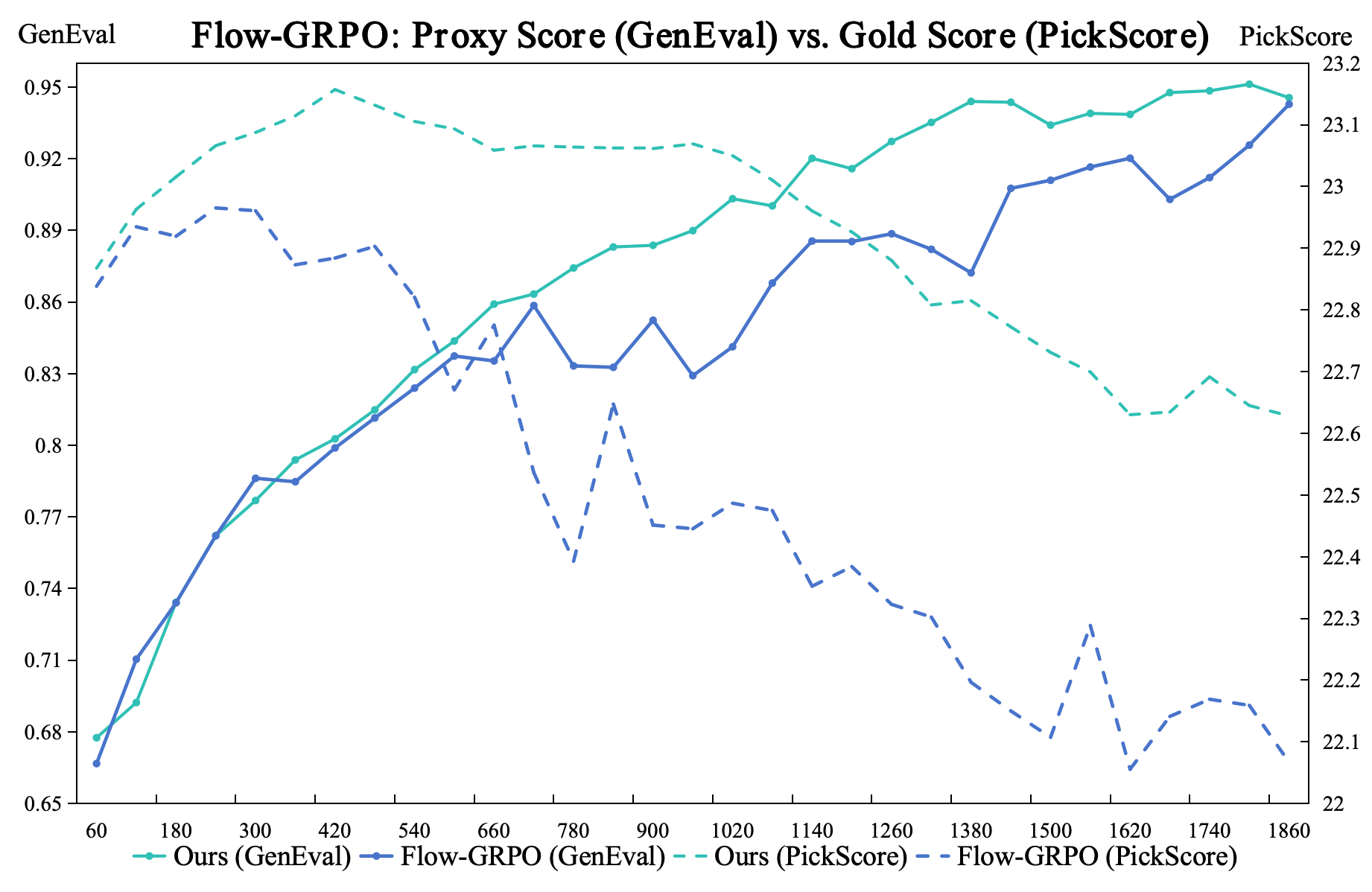}
    \caption{FlowGRPO on the GenEval task.}
    \label{fig:short-a}
  \end{subfigure}
  \hfill
  \begin{subfigure}{0.33\linewidth}
   \includegraphics[width=1.0\linewidth]{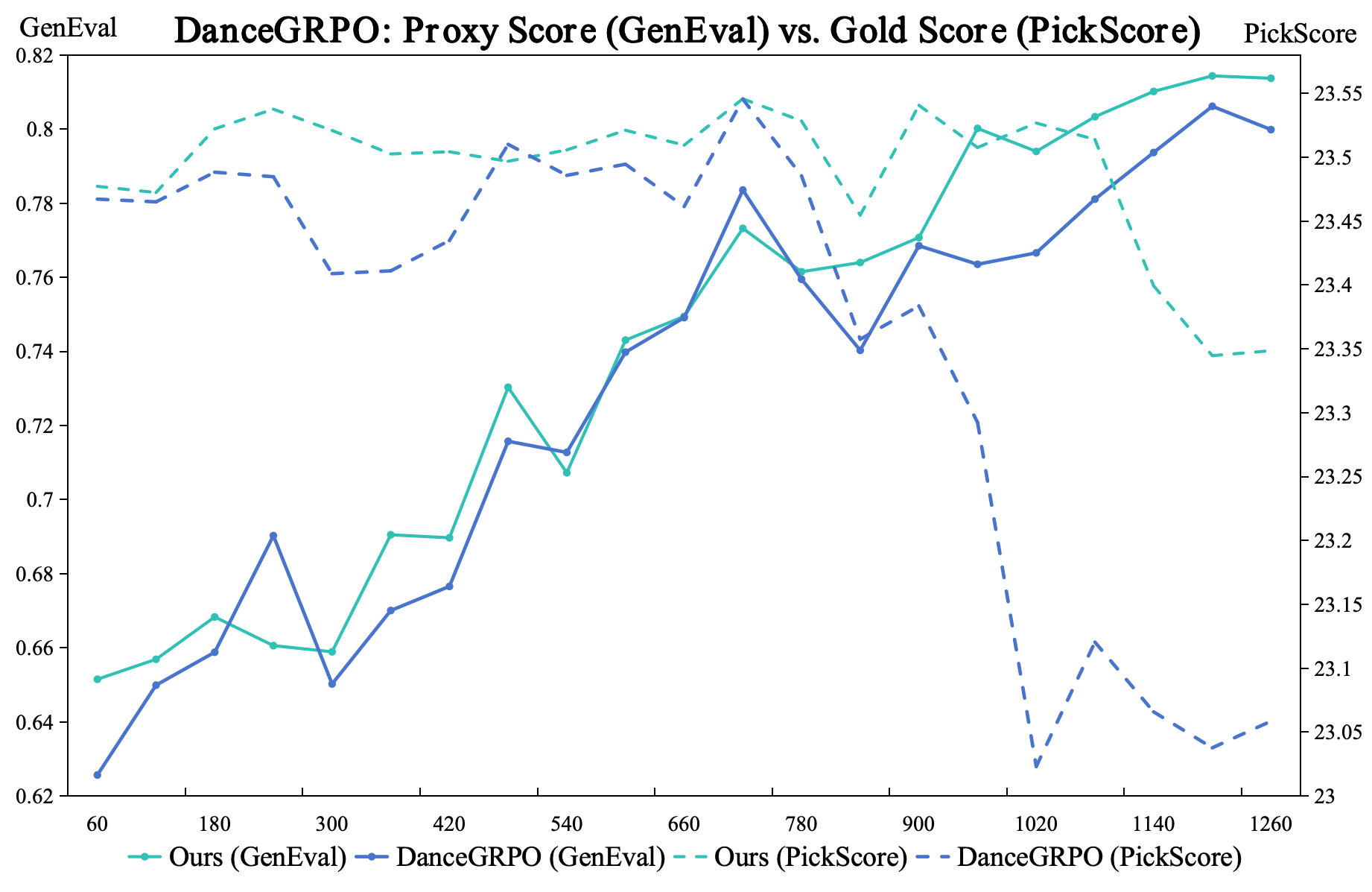}
    \caption{DanceGRPO on the GenEval task.}
    \label{fig:short-b}
  \end{subfigure}
  \hfill
  \begin{subfigure}{0.33\linewidth}
   \includegraphics[width=1.0\linewidth]{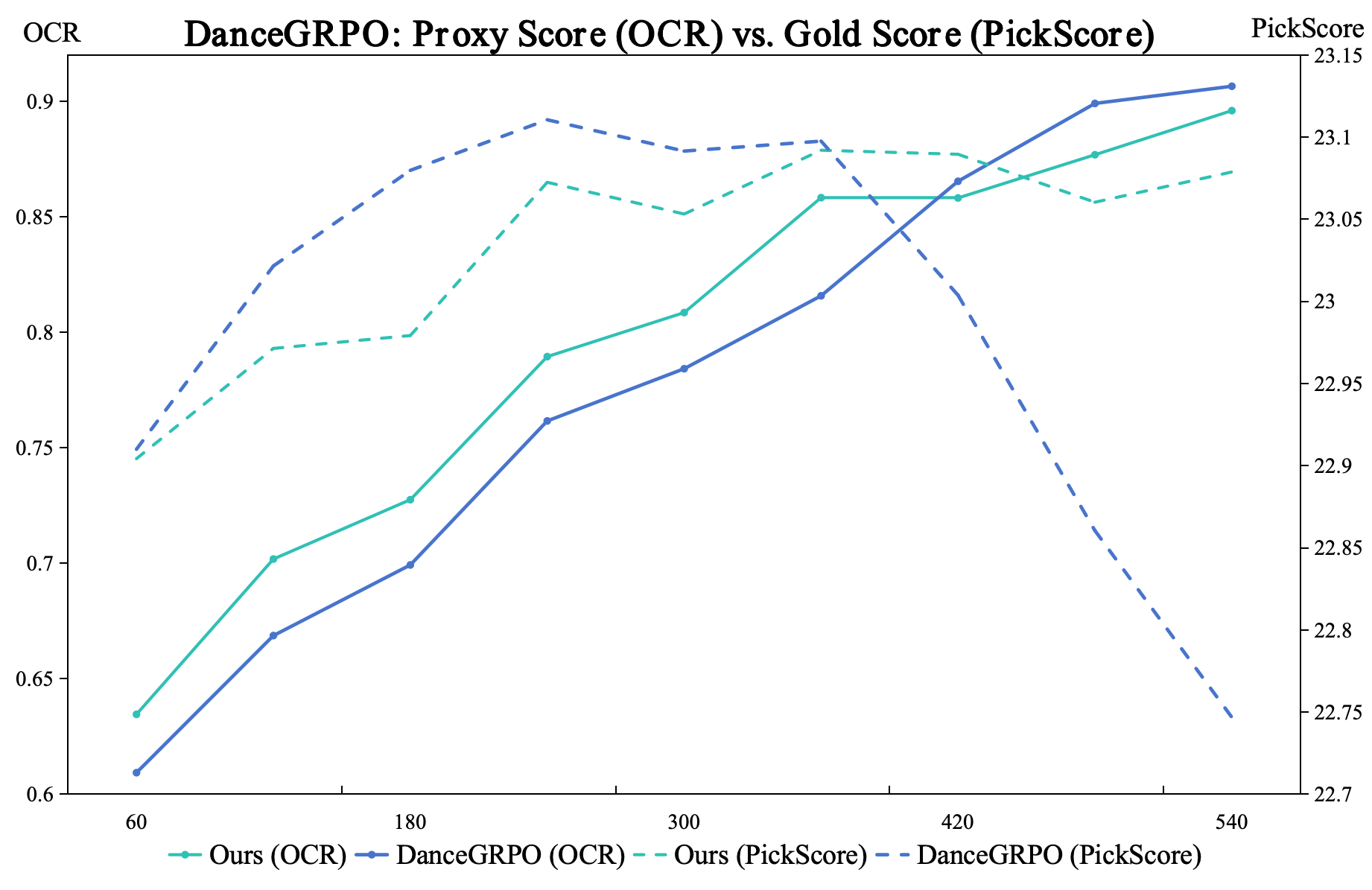}
    \caption{DanceGRPO on the OCR task.}
    \label{fig:short-c}
  \end{subfigure}
  \caption{Validation curves of proxy scores and gold scores across different training tasks and baseline methods.}
  \label{fig:main_training_curve}
\end{figure*}

Combined with the effective clipping mechanism enabled by RatioNorm, GRPO-Guard significantly alleviates the over-optimization phenomenon while maintaining a similar upward trend in the proxy score, as demonstrated in Table \ref{main_result} and Figure ~\ref{fig:main_training_curve}.

\section{Experiments}

\begin{table*}[t]
\caption{Comparison of composite gold scores across different proxy tasks. [·] marks the proxy task associated with each row.
ImR denotes ImageReward, UniR denotes UnifiedReward, and Average represents the mean value after normalizing the three gold scores relative to the base model (set to 1).
}
\vspace{-12pt}
\label{main_result}
\begin{center}
\setlength{\tabcolsep}{1.6mm}
\begin{tabular}{l|l|lll|lllll}
\hline
 \multirow{2}{*}{Method}& \multirow{2}{*}{Step} & \multirow{2}{*}{GenEval} & \multirow{2}{*}{PickScore}  & \multirow{2}{*}{Text Render} & \multicolumn{4}{c}{\bf Gold Score}\\
& & 
& & &\multicolumn{1}{|l}{\bf HPSv2} & \multicolumn{1}{l}{\bf ImR} & \multicolumn{1}{l}{\bf UniR} & \multicolumn{1}{c}{\bf Average} \\ 
\hline
\textcolor{gray}{SD3.5-M~\cite{rombach2022high}}  & - & \textcolor{gray}{0.63}  & \textcolor{gray}{21.5}  & \textcolor{gray}{0.58} & \textcolor{gray}{0.293} & \textcolor{gray}{1.06}  & \textcolor{gray}{3.31} & \textcolor{gray}{1.00}\\
+Flow-GRPO   & 1860 & [0.94]  & 20.4  & 0.59 & 0.236 & 0.85 & 3.05 & 0.84\\
\rowcolor{gray!10}
 +Ours (Flow-GRPO) & 1860  & [0.95]\textcolor{red}{$_{+0.01}$}  & 20.9\textcolor{red}{$_{+0.4}$}  & 0.71\textcolor{red}{$_{+0.12}$} & 0.254\textcolor{red}{$_{+0.018}$} & 0.87\textcolor{red}{$_{+0.02}$} & 3.22\textcolor{red}{$_{+0.17}$} & 0.89\textcolor{red}{$_{+0.05}$} \\
+Flow-GRPO  & 1020 & 0.67  & [23.1]  & 0.64 &   0.329 & 1.40 & 3.46 & 1.16\\
\rowcolor{gray!10}
+Ours (Flow-GRPO) & 1020  & 0.70\textcolor{red}{$_{+0.03}$}  & [23.3]\textcolor{red}{$_{+0.2}$}  & 0.68\textcolor{red}{$_{+0.04}$} & 0.337\textcolor{red}{$_{+0.008}$} & 1.47\textcolor{red}{$_{+0.07}$} & 3.54\textcolor{red}{$_{+0.08}$} & 1.20\textcolor{red}{$_{+0.04}$} \\
+Flow-GRPO   & 480 & 0.52  & 20.8  & [0.94] & 0.274 & 0.82 & 3.07 & 0.88\\
\rowcolor{gray!10}
+Ours (Flow-GRPO) &  480 & 0.65\textcolor{red}{$_{+0.07}$}  & 21.3\textcolor{red}{$_{+0.5}$}  &  [0.93]\textcolor{blue}{$_{-0.01}$} & 0.286\textcolor{red}{$_{+0.012}$} & 1.06\textcolor{red}{$_{+0.24}$} & 3.29\textcolor{red}{$_{+0.22}$} & 0.99\textcolor{red}{$_{+0.11}$} \\
\hline 
\textcolor{gray}{Flux.1-dev~\cite{flux2024}}   & -  & \textcolor{gray}{0.63}  & \textcolor{gray}{21.6} & \textcolor{gray}{0.60} & \textcolor{gray}{0.302} & \textcolor{gray}{1.01} & \textcolor{gray}{3.31} & \textcolor{gray}{1.00}\\
+DanceGRPO  & 1260 & [0.80]  & 21.2  & 0.60 & 0.269 & 0.79 & 3.18 & 0.88\\
\rowcolor{gray!10}
+Ours (DanceGRPO) & 1260  & [0.81]\textcolor{red}{$_{+0.01}$}  & 21.7\textcolor{red}{$_{+0.5}$}  & 0.63\textcolor{red}{$_{+0.03}$} & 0.300\textcolor{red}{$_{+0.031}$} & 1.08\textcolor{red}{$_{+0.29}$} & 3.35\textcolor{red}{$_{+0.17}$} & 1.02\textcolor{red}{$_{+0.14}$}\\
+DanceGRPO & 540  & 0.63  & 21.5  & [0.90] & 0.293 & 0.93 & 3.25 & 0.96\\
\rowcolor{gray!10}
+Ours (DanceGRPO) & 540  & 0.64\textcolor{red}{$_{+0.01}$}  & 21.8\textcolor{red}{$_{+0.3}$}  & [0.89]\textcolor{blue}{$_{-0.01}$} & 0.304\textcolor{red}{$_{+0.009}$} & 1.07\textcolor{red}{$_{+0.14}$} & 3.35\textcolor{red}{$_{+0.10}$} & 1.02\textcolor{red}{$_{+0.06}$}\\
\hline
\end{tabular}
\end{center}
\end{table*}

\subsection{Experimental Setting}

\paragraph{Implementation Details:} We conduct experiments on two baselines, Flow-GRPO~\cite{liu2025flow} and DanceGRPO~\cite{xue2025dancegrpo}, using two backbone models, SD3.5-M and Flux.1-dev, to validate the effectiveness of our method in mitigating reward hacking.
Following the Flow-GRPO setting, we apply LoRA fine-tuning for both baselines, with the LoRA rank set to 32, the scaling factor $\alpha$ set to 64, a learning rate of 3e-4, and a clip range of 1e-4.
For GRPO-Guard, due to the differences in ratio distributions and gradient magnitudes across steps, we set the clip range to 2e-6, with a learning rate of 1e-4 on SD3.5-M and 2e-4 on Flux.1-dev. Notably, since PickScore rewards exhibit relatively minor reward hacking, we use a smaller clip range of 4e-6.
KL loss is not applied. The training and validation datasets are kept consistent with FlowGRPO.

\paragraph{Evaluation Metrics:} Following Flow-GRPO, we conduct experiments on three proxy tasks: GenEval~\cite{ghosh2023geneval}, TextRender~\cite{chen2023textdiffuser}, and PickScore~\cite{kirstain2023pick}.
GenEval is a rule-based evaluation framework that assesses a generator’s ability to follow textual instructions by measuring object count, color consistency, and spatial arrangement.
PickScore is derived from human preference data, where a regression head is fine-tuned on a CLIP encoder so that its scores align with human judgments.
To comprehensively evaluate reward hacking, we further construct a composite gold score based solely on \textbf{image quality}, measured by HPSv2~\cite{wu2023human}, ImageReward~\cite{xu2023imagereward}, and UnifiedReward~\cite{wang2025unified}.
During training, we monitor the gold score online by using PickScore for the GenEval and TextRender tasks. For the validation datasets, GenEval, PickScore, and TextRender use the corresponding validation sets from FlowGRPO, while HPSv2, ImageReward, and UnifiedReward all use the PickScore validation set.

\subsection{Main Results}
We report the results of GRPO-Guard on the GenEval, PickScore, and OCR tasks using two backbone models (Flux and SD3.5M) and two baseline methods (FlowGRPO and DanceGRPO), as shown in Table~\ref{main_result}.
GRPO-Guard achieves superior gold scores under comparable proxy scores, effectively mitigating the severe reward hacking observed in baseline methods caused by the failure of the clipping mechanism.
We further visualize the relationship between proxy scores and gold scores during training in Figure~\ref{fig:main_training_curve}. As training progresses, the proxy scores of baseline models increase rapidly, leading to a sharp decline in gold scores. In contrast, GRPO-Guard maintains consistently high gold scores and image quality throughout the training process.
\begin{figure*}[h]
    \includegraphics[width=1.0\linewidth]{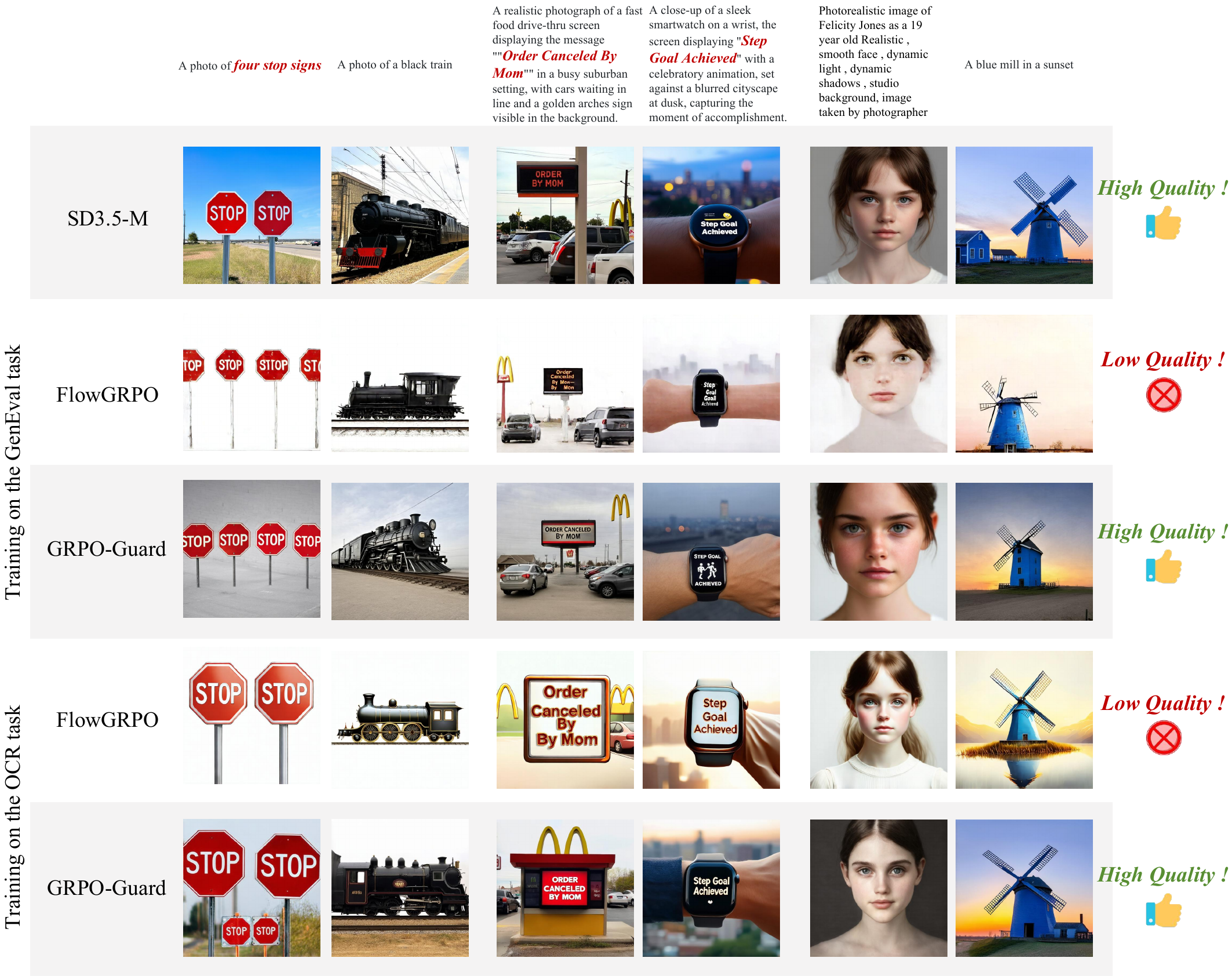}
    \caption{Visual comparison between FlowGRPO and GRPO-Guard. FlowGRPO exhibits clear signs of reward hacking, leading to a significant decline in both image quality and instruction-following ability. In contrast, GRPO-Guard maintains comparable visual quality while demonstrating stronger text generation accuracy and better adherence to instructions.}
    \label{compare_sd35}
\end{figure*}
\begin{figure*}[h]
    \includegraphics[width=1.0\linewidth]{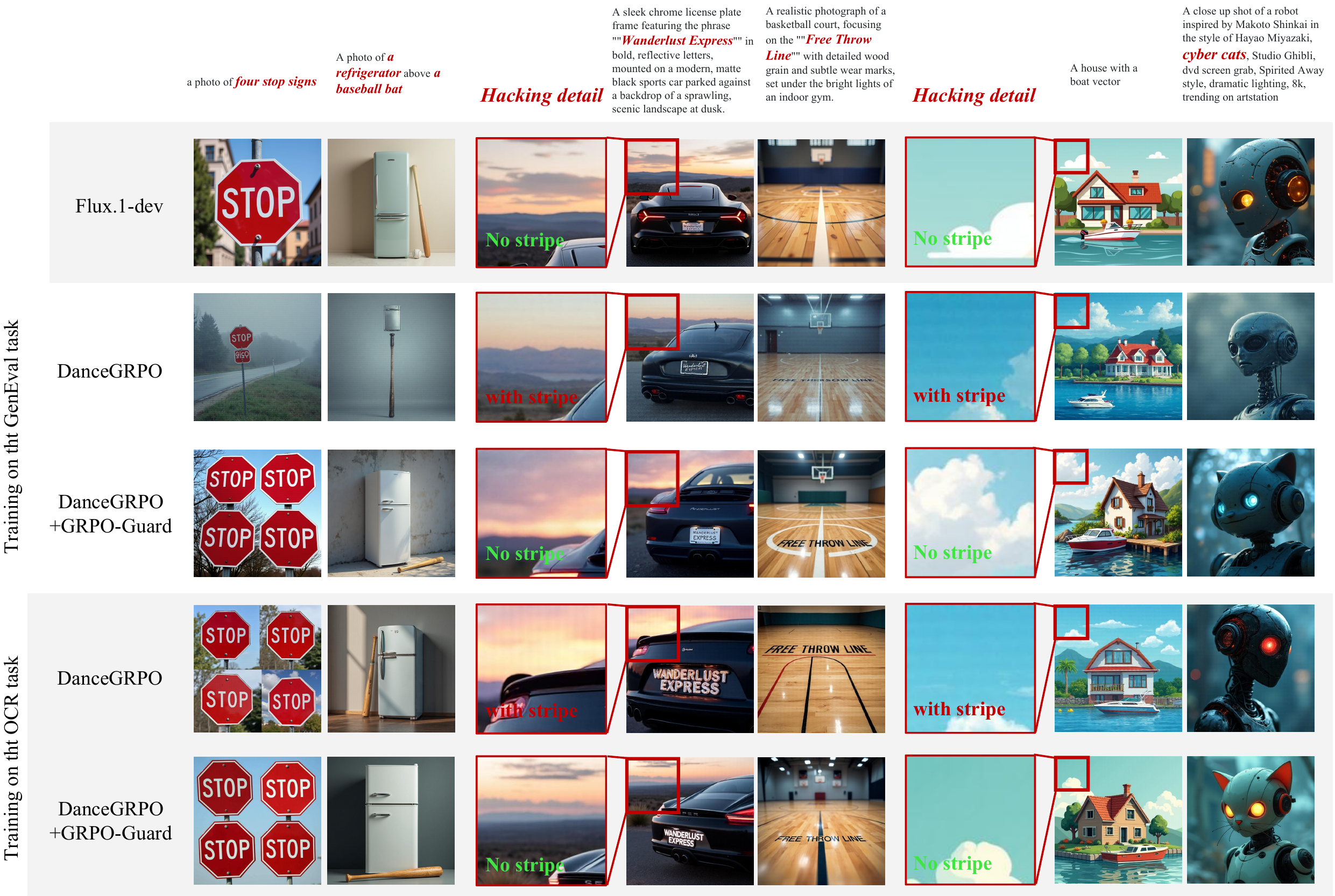}
    \caption{Visual comparison between DanceGRPO and GRPO-Guard. It is clearly observed that DanceGRPO suffers from severe reward hacking, where the generated images exhibit distinct horizontal and vertical stripe artifacts.}
    \label{compare_flux}
\end{figure*}

\begin{figure*}[h]
    \includegraphics[width=1.0\linewidth]{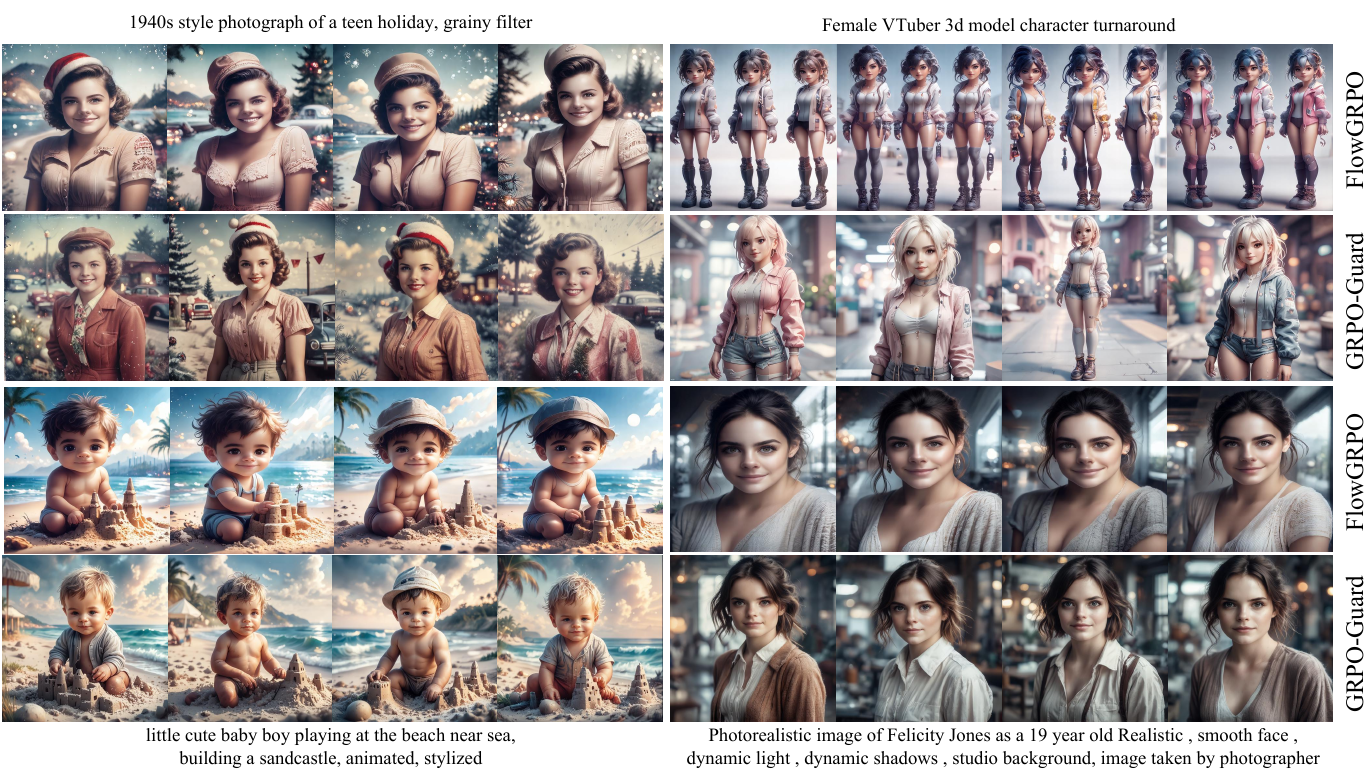}
    \caption{Comparison between FlowGRPO and GRPO-Guard on the PickScore task. FlowGRPO shows severe distortions in human body proportions and a marked reduction in facial diversity, whereas GRPO-Guard effectively preserves realistic body structure and diverse facial appearances throughout training.}
    \label{visual_pickscore}
\end{figure*}

\paragraph{Visual Comparison:} We further provide a visual comparison of the generated images in Figure ~\ref{compare_sd35} and ~\ref{compare_flux}. It is evident that, compared with the original models, the outputs from the baseline methods suffer from severe degradation—the image quality collapses completely. Although these methods achieve high proxy scores, the generated results are unusable in practice.
Notably, while the PickScore results of the baseline methods do not show a significant drop in score, they still exhibit clear reward hacking. As illustrated in Figure~\ref{visual_pickscore}, the generated faces remain nearly identical across different random seeds, and the body proportions become distorted, rendering the outputs impractical for real-world use.
In contrast, our method effectively alleviates these issues while maintaining high proxy scores, producing visually coherent and realistic images.

\begin{figure*}[h]
    \includegraphics[width=1.0\linewidth]{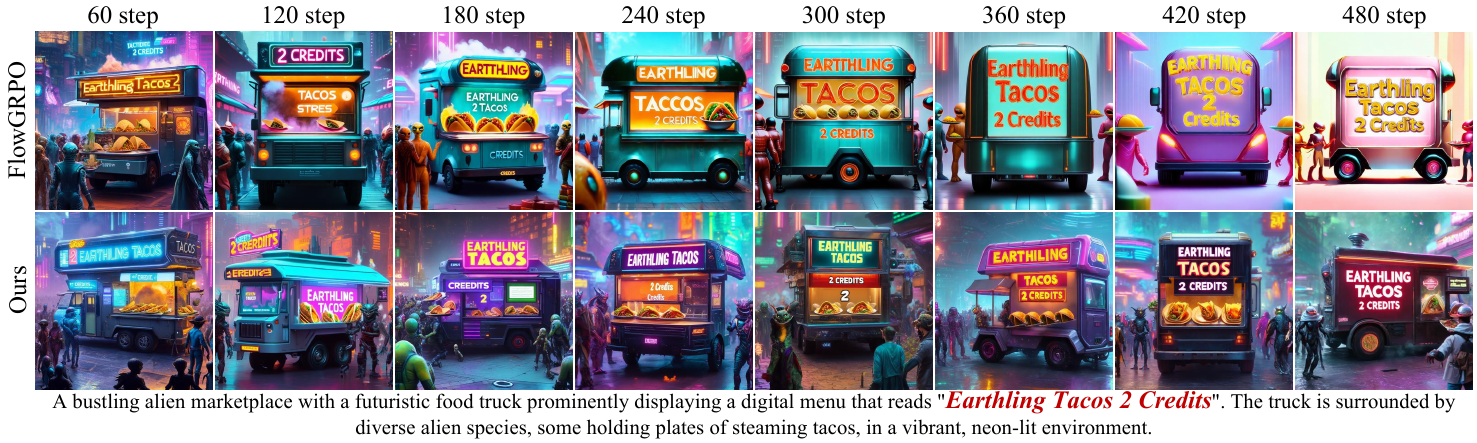}
    \caption{Generation examples of the policy model at different training steps.}
    \label{training_step}
\end{figure*}

\paragraph{Over-optimization:}We further visualize the generated results across different training steps, as shown in Figure~\ref{training_step}. It can be clearly observed that as training progresses, the baseline methods enter an over-optimization phase around the mid-training stage. Due to the failure of the clipping mechanism, the image quality deteriorates rapidly — the proportion of text regions in the generated images increases progressively until complete reward hacking occurs. At this point, the model focuses solely on text correctness, while text-image consistency, scene richness, and diversity collapse entirely. In contrast, our method maintains visual quality comparable to the base model while significantly improving text accuracy, effectively preventing the degeneration observed in baseline methods.

\begin{figure*}[h]
  \centering
  \begin{minipage}[b]{0.60\textwidth}
    \centering
    \small
    \setlength{\tabcolsep}{0.4mm}
    \begin{tabular}{c|c|c|c}
    \hline
    Setting & $\log r_t(\theta)$
    & \makecell{Re-weight \\ Scale} 
    & \makecell{Gradient\\ Scale} 
    \\ \hline
    Baseline & $\log r_t(\theta)$    & 1  & $\beta \frac{\Delta \mu_\theta + \sigma_t \sqrt{dt} \, \epsilon}{\sigma_t^2} $\\
    Temp-Reweight~\cite{he2025tempflow} & $\log r_t(\theta)$  & $\sigma_t \sqrt{dt}$ & $ \beta \frac{\sqrt{dt}\Delta \mu_\theta + \sigma_t dt \, \epsilon}{\sigma_t}$ \\
    Mean-revised & $\log r_t(\theta) + \frac{\|\Delta \mu_\theta\|^2}{2\sigma_t^2 dt}$            & 1 & $\beta \frac{ \sqrt{dt} \, \epsilon}{\sigma_t}$ \\
    RatioNorm & $\sigma_t \sqrt{dt} (\log r_t(\theta) + \frac{\|\Delta \mu_\theta\|^2}{2\sigma_t^2 dt})$             & 1 & $\beta \, dt \,  \epsilon$\\
    GRPO-Guard & $\sigma_t \sqrt{dt} (\log r_t(\theta) + \frac{\|\Delta \mu_\theta\|^2}{2\sigma_t^2 dt})$       & $1/dt$ &  $\beta \, \epsilon $ \\
    \hline
    \end{tabular}
    \captionof{table}{Ablation study on major components.}
    \label{tab:ablation}
  \end{minipage}
  \hfill
  \begin{minipage}[b]{0.38\textwidth}
    \centering
    \includegraphics[width=\linewidth]{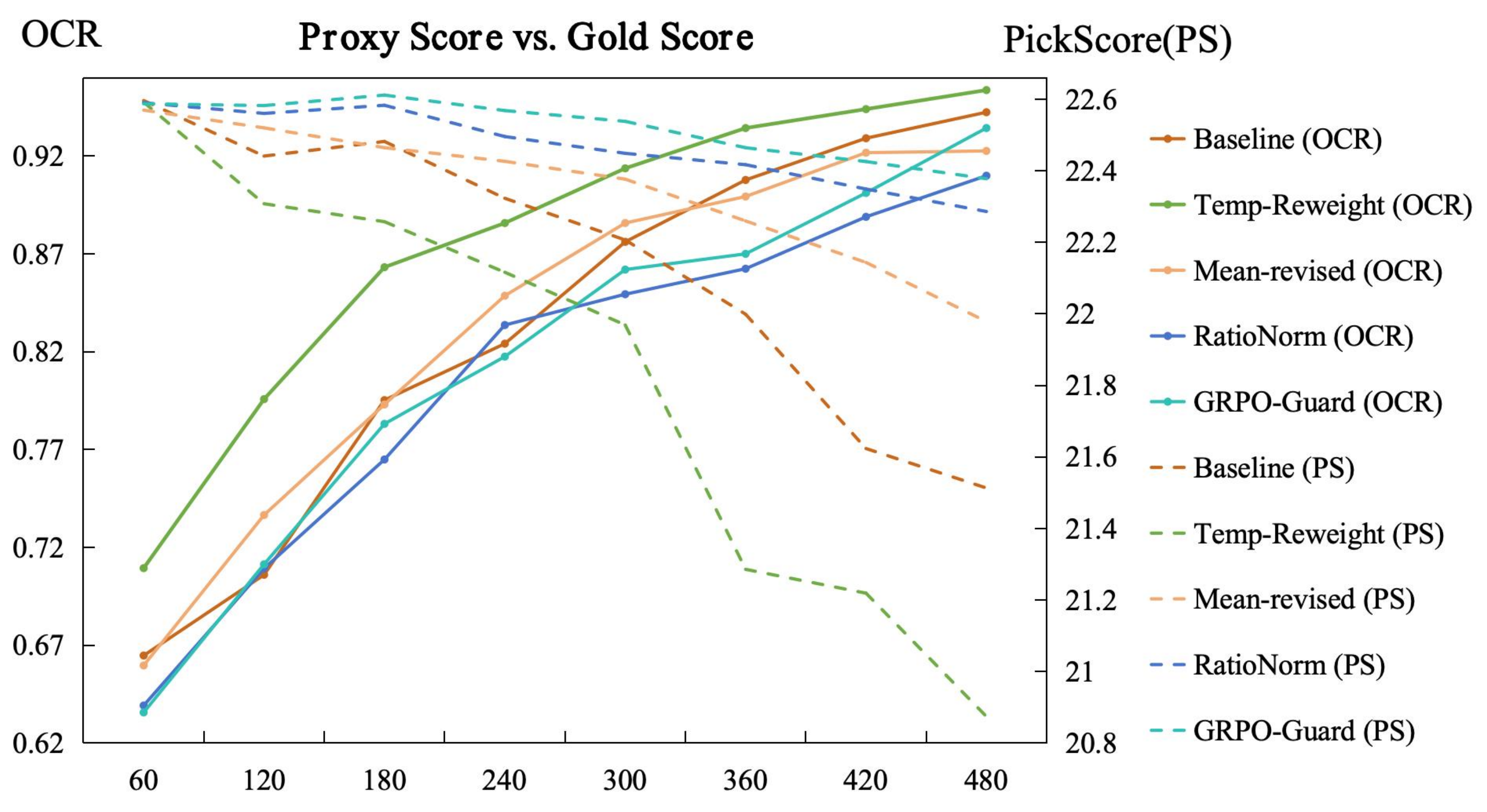}
    \caption{Training curves of the ablation study.}
    \label{fig:ablation}
  \end{minipage}
\end{figure*}

\subsection{Ablation Study}
We further analyze the contributions of the main components of our proposed method. The ablation study is conducted based on the FlowGRPO baseline using the SD3.5-M model on the OCR task, trained for 480 steps. As shown in the Table ~\ref{tab:ablation} and Figure~\ref{fig:ablation}, we design three groups of experiments to separately evaluate the effects of ratio mean correction (Mean-revised), inter-step variance alignment (RatioNorm), and gradient balancing (GRPO-Guard). Their corresponding $\log r_t(\theta)$ distributions and gradient scales are also reported in the table. The experimental results, illustrated in the figure, show that mean correction significantly alleviates the decline in the gold score. Further applying variance alignment mitigates the over-optimization phenomenon even more effectively, although it slightly slows down the growth of the proxy score due to a relatively larger number of positively clipped high-advantage ratios. 
In addition, we compare the reweighting strategy (Temp-Reweight) used in TempFlowGRPO~\cite{he2025tempflow} in terms of optimization efficiency and over-optimization behavior. As shown in the Figure ~\ref{fig:ablation}, although it significantly accelerates optimization, it also enters the over-optimization phase much earlier—resulting in a rapid drop in gold scores and severe reward hacking.
In contrast, the gradient reweighting strategy in GRPO-Guard provides a more moderate improvement in proxy score growth while substantially alleviating the decline in gold scores.

\subsection{Human Evaluation}
\begin{figure}[h]
    \includegraphics[width=1.0\linewidth]{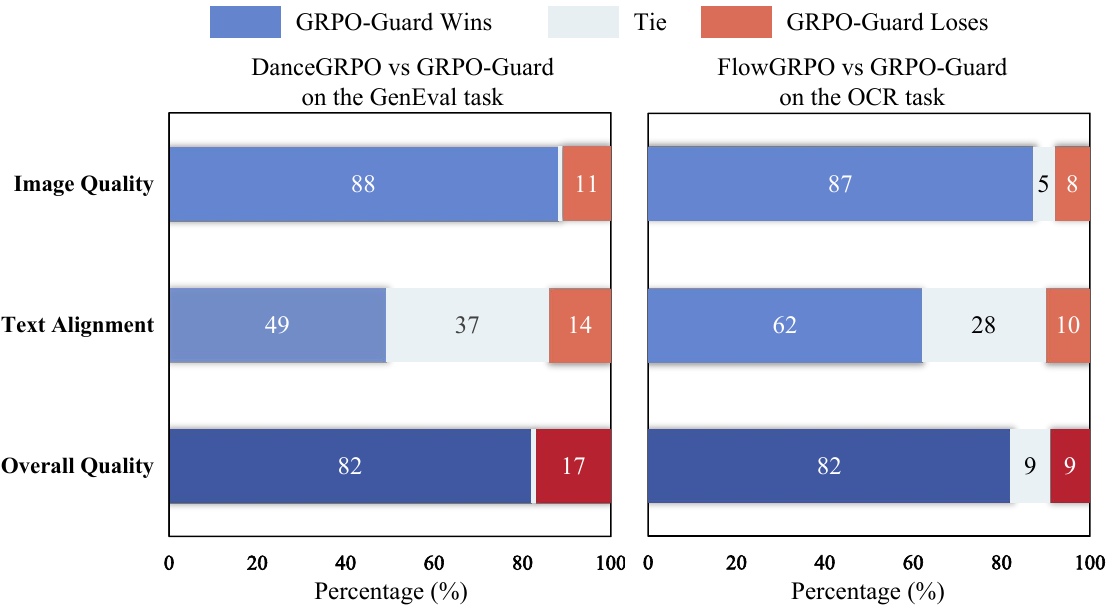}
    \caption{Human evaluation results}
    \label{human_eval}
\end{figure}

We conduct a human preference evaluation to assess image quality, text alignment, and overall quality between the baseline methods and GRPO-Guard. On both the Geneval and OCR tasks, human evaluators compare 100 sample pairs, and the win/tie/lose ratios are shown in Figure~\ref{human_eval}. The results demonstrate a clear superiority of GRPO-Guard in both image quality and overall quality, indicating that the baseline methods suffer from severe over-optimization, leading to a notable degradation in visual fidelity.

\subsection{Analysis}
\begin{figure*}[h]
    \includegraphics[width=1.0\linewidth]{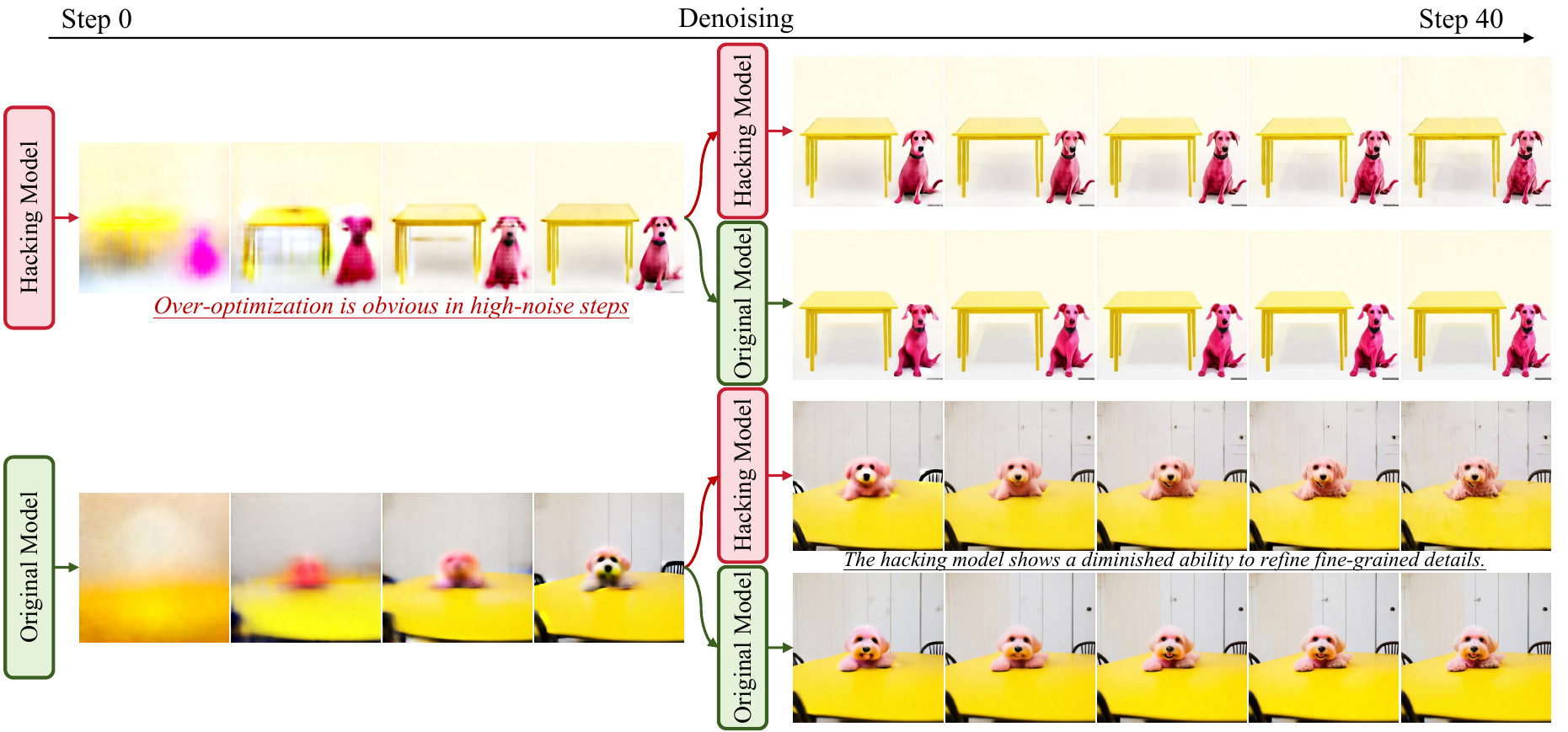}
    \caption{Performance differences between the hacking model and the original model across different denoising steps.}
    \label{hack_step}
\end{figure*}

\paragraph{Hacking Step:}
Due to the malfunctioning clipping mechanism, gradients from all steps with importance ratios exceeding $1 + \epsilon$ are not truncated. Consequently, the hacking model exhibits abnormal behaviors across all denoising stages. we visualize the one-step sampled $x_0$ predictions from $v_\theta$ at different diffusion steps, as shown in Figure~\ref{hack_step}.\textbf{ At high-noise steps}, the hacking model shows clear pathological patterns: the generated images contain overly simplistic and uniform structures—typically limited to the main subjects such as a dog and a table—while omitting broader contextual elements. The global layout appears to be determined prematurely, leaving little room for diverse or detailed scene composition. \textbf{At low-noise steps}, compared with the base model, the hacking model loses its ability to refine fine-grained details. Even during the final denoising stages, substantial residual noise and artifacts remain, resulting in degraded visual quality. These observations indicate that the hacking model suffers from persistent capability degradation throughout the entire denoising process, which aligns with our analysis that gradients beyond $1 + \epsilon$ are never clipped across all timesteps—ultimately causing severe over-optimization.

\begin{figure*}[h]
    \includegraphics[width=1.0\linewidth]{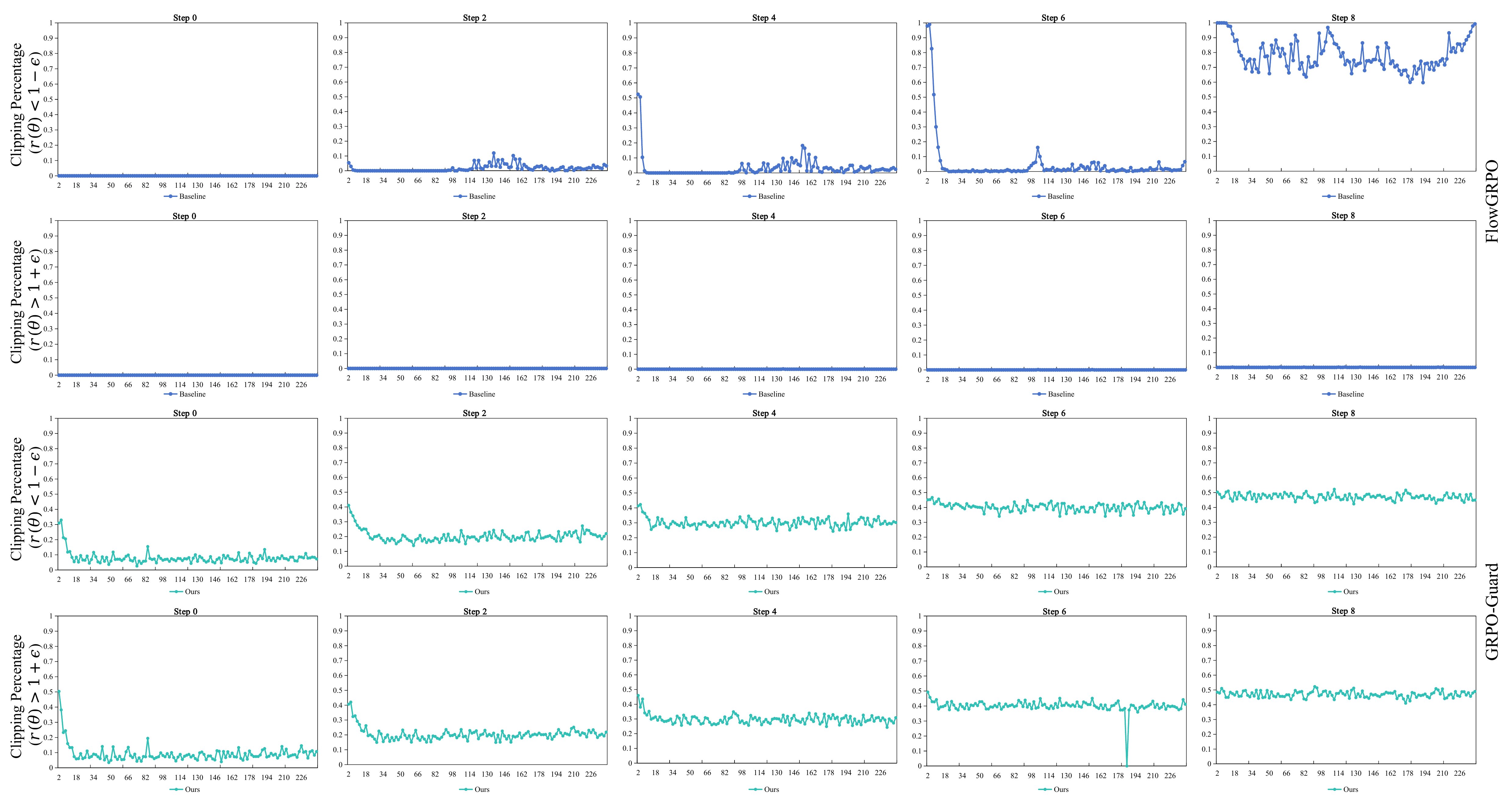}
    \caption{Clipping percentage of $r(\theta) < 1 - \epsilon$ and $r(\theta) > 1 + \epsilon$ during training for FlowGRPO and GRPO-Guard across different denoising steps.}
    \label{clip_frac}
\end{figure*}

\paragraph{Clip Fraction:}
We statistically analyze and visualize the clipping ratios of the baseline methods FlowGRPO and GRPO-Guard across different denoising steps. The proportions of samples with importance ratios $r(\theta)$ larger than $1 + \epsilon$and smaller than $1 - \epsilon$ are recorded separately, as shown in the Figure ~\ref{clip_frac}.
As expected, in FlowGRPO, a large number of clipping events with ratios smaller than $1 - \epsilon$ occur only at the final step (step 8), while the proportion of clipping with ratios larger than $1 + \epsilon$ — corresponding to truncation of gradients with positive advantages — remains zero. This imbalance leads to the over-optimization phenomenon. In contrast, GRPO-Guard exhibits more stable and balanced clipping ratios across all steps, with the proportions of 
$> 1 + \epsilon$ and $< 1 - \epsilon$ clipping remaining roughly equal. This indicates that the distributional bias of the ratio has been effectively corrected and the unhealthy clipping mechanism has been mitigated.

\section{Conclusion and Limitation}

In this paper, we analyze that although GRPO-based reinforcement learning has advanced the optimization of flow-matching models, its standard importance-ratio clipping remains susceptible to over-optimization due to left-shifted and inconsistent ratio distributions. This often leads to deteriorated generation quality despite rising proxy rewards, thereby limiting its practical applicability. GRPO-Guard effectively addresses this issue by incorporating ratio normalization and gradient reweighting, which regulate the clipping mechanism and stabilize policy updates across denoising steps. Extensive experiments demonstrate that GRPO-Guard mitigates over-optimization, preserves or enhances generation quality, and offers a robust, generalizable solution for stable policy optimization. We anticipate that this work will provide valuable insights and practical guidance for the development and optimization of GRPO-based algorithms in flow-matching models.

\paragraph{Limitation:}
Although we effectively mitigate over-optimization by reactivating the clipping capability for positive samples, our approach cannot fully eliminate reward hacking caused by intrinsic limitations of the reward model itself, stemming from the gap between proxy scores (from the reward model) and gold scores (true evaluation). A natural next step to fully address this issue is to scale the reward model, as in approaches like RewardDance~\cite{wu2025rewarddance}, so that it more closely approximates a comprehensive gold score. However, this strategy introduces substantial computational overhead and prolongs optimization, since GRPO requires sampling a large number of outputs along with their reward scores. Therefore, designing a comprehensive, efficient reward model that effectively aligns proxy scores with gold scores remains a promising direction for future research.

\section{ACKNOWLEDGMENTS}
We thank Ziyang Yuan, Borui Liao, Haoran He, Yuanxing Zhang, Qunzhong Wang, Jiaheng Liu for the valuable discussion.


{
    \small
    \bibliographystyle{ieeenat_fullname}
    \bibliography{main}
}


\end{document}